\documentclass{article}

\PassOptionsToPackage{numbers, compress}{natbib}

\usepackage[preprint]{neurips_2026}
\usepackage[utf8]{inputenc}
\usepackage[T1]{fontenc}
\usepackage{hyperref}
\usepackage{url}
\usepackage{microtype}
\usepackage{graphicx}
\usepackage{booktabs}
\usepackage{caption}
\usepackage{subcaption}
\usepackage{amsmath}
\usepackage{amssymb}
\usepackage{algorithm}
\usepackage{algpseudocode}
\usepackage[capitalize,noabbrev]{cleveref}
\usepackage{colortbl}
\usepackage{multirow}
\usepackage{enumitem}
\usepackage{tikz}
\usetikzlibrary{patterns,arrows.meta}

\definecolor{best}{RGB}{250,240,205}   

\definecolor{hlmembg}{RGB}{224,231,255}    
\definecolor{hlmemfg}{RGB}{55,48,163}      
\definecolor{hllatbg}{RGB}{254,243,199}    
\definecolor{hllatfg}{RGB}{146,64,14}      
\definecolor{hltputbg}{RGB}{209,250,229}   
\definecolor{hltputfg}{RGB}{6,95,70}       
\newcommand{\hlmem}[1]{{\setlength{\fboxsep}{1.6pt}\colorbox{hlmembg}{\boldmath\color{hlmemfg}#1}}}
\newcommand{\hllat}[1]{{\setlength{\fboxsep}{1.6pt}\colorbox{hllatbg}{\boldmath\color{hllatfg}#1}}}
\newcommand{\hltput}[1]{{\setlength{\fboxsep}{1.6pt}\colorbox{hltputbg}{\boldmath\color{hltputfg}#1}}}
\definecolor{hllenbg}{RGB}{237,233,254}    
\definecolor{hllenfg}{RGB}{91,33,182}      
\newcommand{\hllen}[1]{{\setlength{\fboxsep}{1.6pt}\colorbox{hllenbg}{\boldmath\color{hllenfg}#1}}}

\definecolor{headcol}{RGB}{243,244,248}   
\newcommand{\hdr}{\bfseries\boldmath}     

\definecolor{cellexact}{RGB}{79,70,181}   
\definecolor{celldrop}{RGB}{203,62,62}    
\definecolor{cellstate}{RGB}{16,155,110}  
\definecolor{okgreen}{RGB}{6,120,86}
\definecolor{badred}{RGB}{180,40,40}

\newcommand{\battn}{\texttt{Attn}}
\newcommand{\bmamba}{\texttt{Mamba}}
\newcommand{\bhybrid}{\texttt{Hybrid}}

\title{Fixed State, Long Reach: What a Constant-Size Cache Buys Block Diffusion at Scale}

\author{\parbox{\textwidth}{\centering
\vspace{0.75cm}
    Vaibhav Singh$^{1,2,3}$\thanks{Correspondence to: vaibhav.singh@mila.quebec} \hspace{-10pt}
    \qquad Pierre-André Noël$^{3}$ 
    \qquad Torsten Scholak$^{3}$ \\
    \qquad Eugene Belilovsky$^{1,2}$
    \qquad Oleksiy Ostapenko$^{3}$
    \vspace{5pt}\\
    \textnormal{{ $^1$Mila \hspace{2pt} $^2$Concordia University \hspace{2pt}$^3$ServiceNow Research}}
}}

\begin{document}
\raggedbottom
\maketitle

\begin{abstract}
Diffusion language models decode tokens in parallel, but their bidirectional denoiser
rules out the naive key--value (KV) cache behind fast autoregressive inference. \emph{Block
diffusion} restores caching by decoding block-by-block, and the block caches deployed on it
so far are tied to \emph{attention}: $\mathcal{O}(L)$ in memory and, if used as training-free
retrofits, only an \emph{approximation} of the model's computation. Both constraints can be overcome: sequence mixers that summarize finalized blocks into a reusable state support block caching, and the corresponding block-causal training objective makes the cache exact. We study this recipe at scale, pretraining three 3B block-diffusion denoisers
(attention, mamba, and hybrid) on $300$B tokens under one single-frontier objective and decoding all three through a single cached interface. Only
the state-space cache is $\mathcal{O}(1)$ in sequence length: its memory and per-step
latency stay \emph{constant} at any context length, while an attention cache remains $\mathcal{O}(L)$. At 256k tokens (where attention has grown to $82$\,GB and
$29$\,ms/step), the Mamba cache delivers \hllat{$4.3\times$} lower latency,
\hlmem{$11\times$} less memory, and \hltput{$2.6\times$} higher single-stream throughput;
and because that footprint is constant it scales with batch as well, reaching
\hltput{$14\times$} the aggregate throughput, where attention cannot run beyond a single
stream. The same linear-state bias lets the Mamba and hybrid backbones keep retrieving out to
\hllen{$8$--$16\times$} their training length, whereas attention's retrieval collapses at
\hllen{$2\times$}, at no measured quality cost.
\end{abstract}

\section{Introduction}
\label{sec:intro}

Diffusion Language Models (DLMs) have emerged as a compelling alternative to the
autoregressive (AR) paradigm that underpins most large language
models~\citep{brown2020language,achiam2023gpt,chowdhery2023palm,touvron2023llama}.
Masked DLMs such as LLaDA~\citep{llada} and Dream~\citep{ye2025dream} corrupt a
sequence by masking tokens and learn to denoise it, decoding many positions in parallel
rather than one token at a time. This promises lower latency and richer per-example supervision~\citep{ni2025diffusion}. In practice, however, DLM
inference has often been \emph{slower} than AR: the very bidirectional attention that
makes denoising possible rules out the naive key--value (KV) cache that AR decoders rely on, so each denoising step recomputes attention over the entire sequence~\citep{fastdllm,ma2025dkv}.

\emph{Block diffusion}~\citep{arriola2025block} resolves the tension by interpolating between the two regimes: the sequence is partitioned into blocks that are generated autoregressively, while diffusion denoising runs \emph{within}
each block. Because finalized blocks form a clean prefix, their representations can be cached and reused, restoring KV-caching and enabling arbitrary-length generation. This has made block diffusion the foundation of recent
fast-decoding systems~\citep{fastdllm,fastdllmv2,wang2025diffusion}.

Two limitations remain. First, the block caches deployed so far are built for an attention backbone~\citep{fastdllm,ma2025dkv,liu2025dllm,nguyen2025elastic};
the resulting KV-cache still grows linearly in context length and the underlying attention still costs quadratic compute, so the long-context regime stays expensive. Second, many of these methods are \emph{training-free} adaptations onto
full-attention models, which makes their block cache an \emph{approximation} of the true bidirectional computation~\citep{fastdllm}. Neither limitation is intrinsic to block diffusion, which only requires that finalized blocks summarize into a reusable state: any sequence mixer with such a state qualifies, and training with the block-causal objective makes the resulting cache exact rather than approximate~\citep{arriola2025block}.

State-space models (SSMs) such as Mamba~\citep{gu2023mamba,dao2024transformers}
are exactly such mixers. They process a sequence with a linear-time recurrence whose hidden state is a fixed-size summary of the past, independent of how long the past is. We build on the bidirectional-Mamba DLM denoiser of DiffuMamba~\citep{singh2026diffumamba} and use its
recurrent state as the block cache: when block diffusion finalizes a block, the forward SSM state already encodes the entire prefix in $\mathcal{O}(1)$ memory. Concurrent work~\citep{chaturvedi2026training} has recently shown, at small scale ($87$M--$350$M parameters), that a Mamba--attention block-diffusion hybrid admits exactly such a cache once the reverse Mamba scan is confined to the active block. Whether this holds when pretrained at scale, how a \emph{pure} Mamba denoiser fares, and what a constant-size state delivers on long-context \emph{retrieval} rather than throughput alone, has not been studied. We therefore expose the SSM state together with the attention KV-cache and a hybrid of the two through a single decoding interface, and we \emph{pretrain} all three models at 3B parameters under the same block-causal objective, so that cached decoding reproduces training exactly.

Concretely, we pretrain and analyze three 3B-parameter block-diffusion models on $300$B tokens that share data, tokenizer, schedule, and decoding budget and differ only in the denoiser: \battn\ (full
attention), \bmamba\ (bidirectional Mamba-2), and \bhybrid\ (attention interleaved every five Mamba layers). Because the Mamba mixer carries larger projections while the MLP is held fixed, \bmamba\ and \bhybrid\ technically have $13\%$ and $11\%$ more parameters than \battn; \cref{sec:res:flops} shows that these extra parameters add a fixed  $\mathcal{O}(d^2)$ cost per token at every context length, whereas attention's score computation costs $\mathcal{O}(Ld)$ per token and grows with context. Per-token FLOPs are comparable at the training length, and the long-context gap is due to the mixer rather than the parameter budget. Our contributions are:

\begin{itemize}[leftmargin=1.4em,itemsep=3pt,topsep=2pt]
\item \textbf{A controlled 3B pretraining study of cacheable block-diffusion denoisers}
  (\cref{sec:method,sec:exp}): attention, bidirectional Mamba-2, and hybrid
  backbones, each pretrained on $300$B tokens under the same single-frontier
  block-causal objective. As in BD3LM \citep{arriola2025block} for attention
  and, concurrently, \citet{chaturvedi2026training} for Mamba hybrids, the cache is
  exact because the models are trained with the objective used at decode time;
  unlike training-free retrofits, cached inference \emph{is} the function the model
  learned. To our knowledge this is the largest such study, and the only one that
  includes a \emph{pure} state-space denoiser.
\item \textbf{Constant memory and long context from a single linear-state backbone}
  (\cref{sec:res:eff,sec:res:long}). Measured on the trained checkpoints, the SSM
  cache is $\mathcal{O}(1)$ in length: \bmamba\ holds $\sim$7.5\,GB and 6.8\,ms/step
  at \emph{every} length out to 256k, against attention's 82\,GB, reaching
  \hltput{$\mathbf{14\times}$} the aggregate throughput at 256k. The same backbone
  also generalizes far past its 1024-token training length: \battn's retrieval
  collapses just $2\times$ out on NIAH and LongBench, whereas \bmamba/\bhybrid\
  extrapolate to \hllen{$8$--$16\times$}.
\item \textbf{Long-context evaluation at scale} (\cref{sec:results}). With everything
  but the denoiser held fixed, we cover cached-decoding efficiency (latency, memory,
  and throughput) out to 256k tokens, NIAH and LongBench
  length extrapolation, an eight-task downstream suite, MAUVE and generative
  perplexity, and a per-layer FLOPs analysis that attributes the gains to the
  \emph{architecture} rather than to parameter count.
\end{itemize}

None of these properties is new to state-space models. Fixed-size state~\citep{gu2023mamba,dao2024transformers} and length
extrapolation~\citep{de2024griffin,ren2024samba} are established properties of \emph{autoregressive} SSMs, with hybrids restoring the recall that pure recurrence lacks~\citep{waleffe2024empirical,lieber2024jamba}. Whether they carry over to diffusion denoisers, which decode several tokens per pass, is open. Our results are the block-diffusion
counterpart of that AR literature, and the combination matters because block diffusion reveals
several tokens per forward pass rather than one, and a linear-state backbone lets it do so on a
state that never grows. \citet{singh2026diffumamba} showed
Mamba denoisers match attention on quality at the training length, and \citet{chaturvedi2026training} that they can be block-cached like attention~\citep{arriola2025block} at small scale; we show that the same backbones, pretrained at 3B,
\emph{decode} long contexts with a constant-memory cache and retrieve far past their training length, extending the case for linear-state diffusion LMs from training to inference and from
throughput to long-context retrieval.

\section{Background and Related Work}
\label{sec:related}

\paragraph{Masked diffusion language models.}
Discrete diffusion models define a forward process that progressively corrupts a
token sequence and learn a parameterized reverse process that
denoises~\citep{d3pm,campbell2022continuous}. The \emph{masked} (absorbing-state)
instantiation has proven the most effective for language: SEDD~\citep{lou2023discrete},
MDLM~\citep{sahoo2024simple}, and their refinements~\citep{shi2024simplified,duo}
train a bidirectional network to recover masked tokens under a noise schedule,
and LLaDA~\citep{llada} and Dream~\citep{ye2025dream} scale this recipe to
billions of parameters. All decode by iteratively unmasking positions in
parallel; none admits a KV-cache natively, because every denoising step attends
over the full (partially masked) sequence in both directions.

\paragraph{Block diffusion.}
BD3LM~\citep{arriola2025block} interpolates between AR and diffusion: tokens are
grouped into blocks generated left-to-right, with intra-block diffusion. Blocks
already produced act as a clean prefix, so their keys and values can be cached
between blocks, and sequences of arbitrary length can be generated. Our attention
model follows this scheme; we use the name \battn\ rather than ``BD3LM'' to keep
our trained model distinct from the cited training recipe. Subsequent systems
build fast decoders on this foundation~\citep{fastdllm,fastdllmv2,wang2025diffusion}.

\begin{figure}[t]
\centering
\begin{subfigure}{0.8\linewidth}
  \centering
  \includegraphics[width=\linewidth]{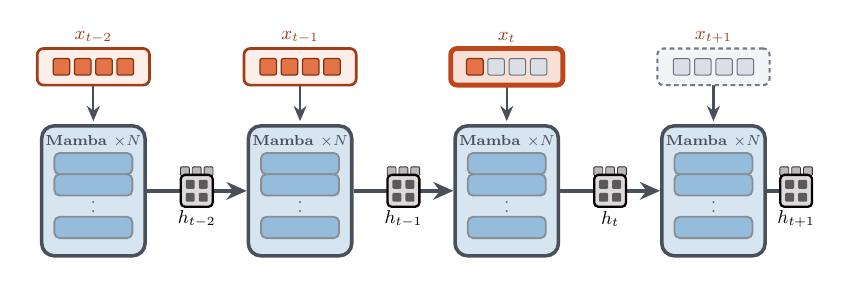}
  \caption{\small \bmamba/\bhybrid: a single recurrent state $h$ of fixed shape
$\mathbb{R}^{H\times P\times N}$, carried and updated in place
($h_t=\bar A_t\,h_{t-1}+\bar B_t\,x_t$), $\mathcal{O}(1)$ in sequence length.}
  \label{fig:method:mamba}
\end{subfigure}

\begin{subfigure}{0.8\linewidth}
  \centering
  \includegraphics[width=\linewidth]{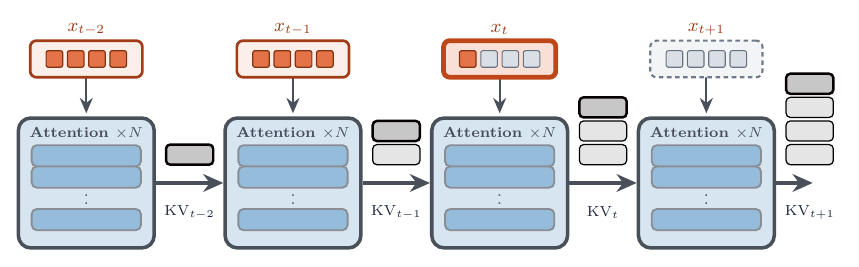}
  \caption{\small \battn: the key--value memory, which appends one entry per
block ($\mathrm{KV}_{t-2}\!\to\!\mathrm{KV}_{t+1}$) and therefore grows as
$\mathcal{O}(L)$.}
  \label{fig:method:attn}
\end{subfigure}
\caption{\small \textbf{The unified block cache.} Block diffusion decodes one block at a
time, left to right: finalized blocks ($x_{t-2},x_{t-1}$) are clean, the frontier
block $x_t$ is being denoised (its first token is revealed, the rest are
\texttt{[MASK]}), and future blocks ($x_{t+1}$) stay fully masked. Once a block is
finalized, every layer writes its state into a cache that the next block reads, so the
clean prefix is never recomputed. The cached object depends on the backbone.}
\label{fig:method}
\vspace{-10pt}
\end{figure}
\paragraph{Caching for diffusion LMs.}
Because the bidirectional denoiser blocks the standard KV-cache, a line of work
restores caching for attention DLMs. Fast-dLLM~\citep{fastdllm} introduces a
block-wise \emph{approximate} KV-cache plus confidence-aware parallel decoding,
reporting large speedups on LLaDA/Dream by caching the fixed context and
refreshing it at block boundaries; dKV-Cache~\citep{ma2025dkv} proposes a delayed
KV-cache compatible with bidirectional attention; and dLLM-Cache~\citep{liu2025dllm}
and elastic-cache variants~\citep{nguyen2025elastic} add adaptive eviction and
reuse. A more recent wave targets the \emph{long-context} regime specifically:
Prefilling-dLLM~\citep{xiong2026prefilling} caches chunked prefix KV and keeps the
top-$K$ relevant chunks, and Focus-dLLM~\citep{long2026focusdllm} exploits attention
sparsity for dynamic cache eviction, reporting $9$--$28\times$ speedups at $8$--$32$k.
A separate line extends the \emph{usable} context of attention DLMs by rescaling
positional encodings: LongLLaDA~\citep{liu2025longllada} applies training-free
NTK-RoPE (and documents a ``local perception'' sliding-window bias), while
UltraLLaDA~\citep{he2025ultrallada} \emph{trains} to a $128$k window. All of these
caches are attention-only and $\mathcal{O}(L)$ in memory, and, being largely
training-free retrofits, only \emph{approximate} the full-attention computation
they replace. The exception is training-time block-causal schemes, BD3LM~\citep{arriola2025block}
for attention and, concurrently with this work, \citet{chaturvedi2026training} for
Mamba--attention hybrids, whose caches are exact because the model is trained on the same
clean-prefix computation it decodes with; our models belong to this family.

\paragraph{State-space models and hybrids.}
SSMs model sequences with a linear-time selective recurrence: S4~\citep{gu2021s4},
Mamba~\citep{gu2023mamba}, and Mamba-2~\citep{dao2024transformers} match or beat
attention on language while carrying a fixed-size state, and related structured
operators~\citep{poli2023hyena,dao2023h3} share the linear-time property. Hybrid
AR models interleave attention with linear recurrence to combine global recall
with cheap long-range mixing~\citep{lieber2024jamba,de2024griffin,wang2025systematic}.
DiffuMamba~\citep{singh2026diffumamba} brought bidirectional Mamba and Mamba--attention
hybrids to \emph{diffusion} denoisers. \citet{chaturvedi2026training} trained block cache diffusion models at a small scale with
Mamba--attention hybrids of $87$M and $350$M parameters. 

What this study adds is scale and evidence: pretraining
at 3B parameters and $300$B tokens with all three backbones, including a \emph{pure}
Mamba denoiser absent from that work; a single-frontier objective; latency, memory,
and throughput measured on the trained checkpoints, single-stream and batched; and,
most importantly, long-context \emph{retrieval} and length extrapolation (NIAH,
LongBench) alongside downstream and generation quality. Our claim is therefore not
the mechanism but its behavior at scale: constant-memory block decoding survives
pretraining at 3B and comes with a marked long-context retrieval advantage over the
attention baseline.

\section{Method}
\label{sec:method}

\begin{algorithm}[t]
\caption{Cached block-wise generation}
\label{alg:cached-gen}
\begin{algorithmic}[1]
\Require prefix, block size $G$, steps per block $S$, blocks $K$, mask id
\State $\mathit{cache} \gets \varnothing$;\quad warm $\mathit{cache}$ on the clean prompt blocks
\For{$i \gets i_{\text{first}}$ \textbf{to} $K-1$}
  \State $s \gets \textsc{Clone}(\mathit{cache})$ \Comment{entry state, reused across denoising steps}
  \State $m \gets$ \# masked tokens in block $i$
  \For{$t \gets 1$ \textbf{to} $S$}
    \State $\ell \gets \textsc{ForwardBlockCached}(x_i,\, s,\, \mathrm{pos}_i)$
    \State reveal the $\lceil m/S\rceil$ highest-confidence masked positions of $x_i$ using $\ell$
    \If{block $i$ fully revealed} \State \textbf{break} \EndIf
  \EndFor
  \State $\_,\ \mathit{cache} \gets \textsc{ForwardBlockCached}(x_i,\, s,\, \mathrm{pos}_i)$ \Comment{fold finalized block into cache}
\EndFor
\State \Return $x$
\end{algorithmic}
\end{algorithm}

\subsection{Block-diffusion preliminaries}
\label{sec:method:prelim}
Let $x=(x_1,\dots,x_L)$ be a token sequence. A masked diffusion model defines a
forward process that, at time $t\in(0,1]$, independently replaces each token by a
special \texttt{[MASK]} symbol with probability $1-\alpha_t$, where $\alpha_t$
follows a monotone schedule. The denoiser $f_\theta$ is trained to predict the
original tokens from the masked sequence, minimizing a re-weighted cross-entropy
that forms a negative ELBO on the data log-likelihood~\citep{sahoo2024simple,llada}.

\emph{Block diffusion} factorizes the sequence into $K=L/G$ contiguous blocks of
size $G$ and is autoregressive across blocks: block $i$ is generated conditioned
on the clean blocks $0,\dots,i{-}1$, while the tokens \emph{inside} block $i$ are
produced by diffusion. We train with a \textbf{single-frontier} objective
(\cref{fig:method}): for
each example we sample one frontier block $i$, keep blocks $<i$ clean, mask a
time-dependent random subset of block $i$, and mask blocks $>i$ \emph{entirely};
the loss is the re-weighted cross-entropy on the masked positions of block $i$
only. This exactly matches the information available during cached generation,
where block $i$ sees a clean prefix and all-masked future. The frontier structure is enforced differently per backbone.
\emph{Attention} layers use a \textbf{block-causal} mask: a query in block $i$
attends to all keys in blocks $\le i$ (bidirectionally within its own block) and
to none in blocks $>i$. \emph{Mamba} layers run a forward (left-to-right)
recurrence over the whole prefix, and a backward (right-to-left) recurrence whose
support is \emph{restricted to the frontier block}, so that no information leaks
backward from the masked future (the block-restricted reverse scan that
\citet{chaturvedi2026training} term \emph{partial bidirectionality}). The hybrid
applies both rules, layer by layer.

\subsection{A unified block cache}
\label{sec:method:cache}
At inference the blocks are produced strictly left-to-right, so once block $i$ is
finalized its contribution to every future block is fixed. We expose a single
per-layer interface that consumes only the $G$ tokens of the current
block plus a per-layer cache summarizing blocks $0,\dots,i{-}1$, and returns the
block logits and an updated cache. The cache type is backbone-specific
(\cref{tab:configs}). In the attention layers, for the $G$ new tokens, we compute queries, keys, and values, and then
concatenate the new keys/values with the cached $(K,V)$ of all finalized blocks. The attended context is the clean prefix plus the
current block; the cache grows as $\mathcal{O}(L)$ and the per-step cost is
$\mathcal{O}(GL)$.

A Mamba-2 layer maintains two states that together summarize the prefix: a short
causal-convolution state (the last $k{-}1$ inputs, where $k$ is the conv width)
and the selective-SSM recurrent state of shape
$(\text{heads}\times\text{head\_dim}\times d_{\text{state}})$. The forward
recurrence consumes the cached states, emits the block output, and returns
updated states via a chunked scan with initial states; both states are
\emph{independent of $L$}. The backward (bidirectional) recurrence is re-run from
a zero state \emph{within the current block only}, reproducing the frontier
restriction used in training; it is never carried across blocks. Thus the
Mamba cache is $\mathcal{O}(1)$ in length and the per-step cost is
$\mathcal{O}(G^2)$, independent of how much text precedes the block. Lastly the hybrid denoiser interleaves the two layer types, so its cache is a heterogeneous list: attention layers store $(K,V)$ tensors and Mamba layers store
$(\text{conv},\text{ssm})$ states. The
overall memory is dominated by the (few) attention layers, giving a cost between
those of the two pure backbones.

\begin{figure}[t]
\centering
\includegraphics[width=\textwidth]{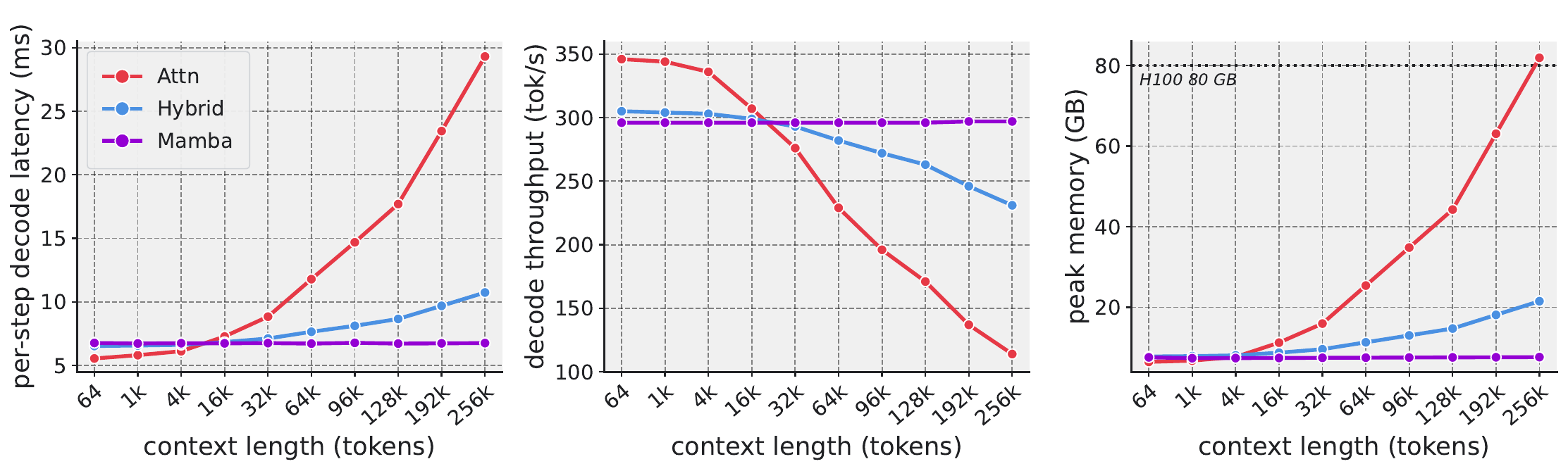}
\caption{\textbf{Cached-decoding efficiency vs.\ context length (3B, single H100 80\,GB,
\texttt{bf16}, cudagraph, batch~1).} (a)~per-step decode latency, (b)~end-to-end
decode throughput ($S{=}16$), (c)~peak memory. \bmamba\ is flat at every
length to 256k (constant SSM state), while \battn\ grows as $\mathcal{O}(L)$: by 256k
\bmamba\ is $\mathbf{2.6\times}$ faster (296 vs.\ 114\,tok/s) at $\mathbf{11\times}$ less
memory (7.5 vs.\ 82\,GB); \bhybrid\ interpolates. \cref{fig:efficiency_batch} sweeps
batch $1$--$8$.}
\label{fig:efficiency}
\vspace{-15pt}
\end{figure}

\subsection{Inference via cached generation}
\label{sec:method:gen}
Generation proceeds block by block (\cref{alg:cached-gen}). Entering block $i$, we
clone the layer cache so its entry state can be reused across the block's denoising
steps (each cached forward returns fresh tensors and never mutates its inputs). We
then run $S$ denoising steps: at each step a cached forward pass scores the current
block, and we reveal the $\lceil m/S\rceil$ highest-confidence still-masked
positions, where $m$ is the number of masked tokens at block entry. Once the block
is fully revealed, one final cached pass folds it into the cache that block $i{+}1$
reads. With $S$ the only quality--speed knob, the total number of forward passes is
$\mathcal{O}(KS)=\mathcal{O}((L/G)S)$, and for the SSM backbone each pass touches
only the $G$ current tokens regardless of $L$.

\subsection{Complexity and asymptotic throughput}
\label{sec:method:complexity}
\cref{tab:configs} summarizes the cost of one cached
denoising step and the state it carries. Decoding the frontier block of $G$ tokens against a prefix of
length $L$ costs $\mathcal{O}(GL)$ for attention -- each new query attends over
the growing KV-cache -- so its per-step latency grows linearly in $L$ and the
decode throughput scales as $T=\mathcal{O}(1/L)$. The Mamba model consumes a
fixed-size SSM state and touches only the $G$ current tokens, giving an
$L$-independent per-step cost of $\mathcal{O}(G^2)$ and hence constant
throughput. The hybrid inherits the attention scaling with a small constant
($5/28$ of its layers carry attention). \cref{sec:res:eff} measures exactly these
three curves.

\section{Experimental Setup}
\label{sec:exp}

\paragraph{Models.}
We compare three 3B-parameter denoisers that are identical except for the
sequence mixer (\cref{tab:configs}): \battn\ (28 attention layers), \bmamba\ (28
bidirectional Mamba-2 layers), and \bhybrid\ (5 attention layers interleaved with
23 Mamba layers, following \citet{singh2026diffumamba}). All models use $d_{\text{model}}=2560$, block size $G=32$ and are trained on Nemotron-CC~\citep{su2024nemotron} with $300$B token budget under the single-frontier block-diffusion objective (\cref{sec:method:prelim}) which masks a random subset of a single sampled frontier block per example, keeping earlier blocks clean and later blocks fully masked, so training sees exactly the clean-prefix, all-masked-future context each block meets at cached decode time. We trained with sequence length of $1024$ tokens, and the global batch being $4096$ sequences ($1024\times4096\approx4.2$M tokens per step) with a $71\,525$-step schedule covering $1024\times4096\times71\,525\approx300$B tokens. Optimization is AdamW in \texttt{bf16} (learning rate $1\times10^{-4}$, cosine schedule, $5\%$ warmup, weight decay $0.1$, gradient-norm clip $1.0$); the complete set of training and model hyperparameters is in \cref{app:hparams}. 

\begin{table}[t]
\centering
\caption{\textbf{Three denoisers, one variable: the mixer.} The 3B models share $d_{\text{model}}$, depth, block size $G$, and
differ only in the sequence mixer. The right block gives each backbone's per-step decode cost over a block of $G$ tokens against a prefix of length $L$: only the SSM
cache is $\mathcal{O}(1)$ in $L$ (conv+SSM state vs.\ an $\mathcal{O}(L)$ KV-cache), the root of \bmamba's flat memory and latency; $^{\ast}$small constant ($5/28$ layers carry attention).}
\label{tab:configs}
\small
\setlength{\tabcolsep}{4pt}
\resizebox{\linewidth}{!}{%
\begin{tabular}{lcccccc ccc}
\toprule
\rowcolor{headcol}
& \multicolumn{6}{c}{\hdr Configuration} & \multicolumn{3}{c}{\hdr Per-step cost (block $G$, prefix $L$)} \\
\cmidrule(lr){2-7}\cmidrule(lr){8-10}
\rowcolor{headcol}
\hdr Model & \hdr $d_{\text{model}}$ & \hdr Layers & \hdr Mixer & \hdr $G$ & \hdr Train ctx.\ & \hdr best val.\ NLL & \hdr Compute & \hdr Cache mem & \hdr Throughput \\
\midrule
\battn   & 2560 & 28 & 28 attn            & 32 & 1024 & 1.104 & $\mathcal{O}(GL)$ & $\mathcal{O}(L)$ & $\mathcal{O}(1/L)$ \\
\bhybrid & 2560 & 28 & 5 attn + 23 Mamba  & 32 & 1024 &  \cellcolor{best}1.102 & $\mathcal{O}(GL)^{\ast}$ & $\mathcal{O}(L)^{\ast}$ & $\mathcal{O}(1/L)^{\ast}$ \\
\bmamba  & 2560 & 28 & 28 bidir.\ Mamba-2 & 32 & 1024 & 1.134 & $\mathcal{O}(G^2)$ & $\mathcal{O}(1)$ & $\mathcal{O}(1)$ \\
\bottomrule
\end{tabular}%
}
\vspace{-10pt}
\end{table}
\vspace{-10pt}
\paragraph{Hardware and timing.}
All efficiency measurements use a single NVIDIA H100 80\,GB in \texttt{bf16} with
PyTorch SDPA attention kernels. We time one denoising step by CUDA-graph
capture which removes kernel launch overhead, and report the mean of 50
replays under CUDA events. End-to-end decode throughput is the ratio of total tokens generated to the total time to decode them, adding up each block's latency at its true cache depth (block $i$ is decoded against a prefix of
length $i\cdot G$).
\vspace{-10pt}
\paragraph{Evaluation protocols.} We measured \emph{Efficiency} by sweeping context length $L$ from $64$ to $256$k tokens at batch sizes
$1$--$8$, and report per-step decode latency, peak memory, and end-to-end decode
throughput at $S{=}16$ denoising steps per block.
For \emph{Long Context Performance} we measure retrieval with a RULER-style~\citep{hsieh2024ruler}
needle-in-a-haystack (NIAH) passkey task, where a short key is hidden at one of five
depths inside a long distractor context and must be recovered ($20$ trials per
(length, depth), $L$ from $512$ to $16$k). We further evaluated the models on
LongBench~\citep{bai2024longbench}, a $16$-task English-and-code long-context suite
(middle-truncated to $L$, $20$ examples per task, $L$ from $2$k to $16$k). For the attention backbones we additionally report an NTK-RoPE
variant~\citep{peng2024yarn,liu2025longllada}: a training-free rescaling of RoPE's
rotation frequencies that folds the longer positions seen at inference back toward the
range covered during training, extending the usable window without any fine-tuning. For
\emph{Downstream Performance} we score eight common-sense / reasoning tasks with the
block-diffusion likelihood harness, and for evaluating
\emph{Generation Quality} we report
generative perplexity (scored by GPT-2 large) alongside MAUVE~\citep{pillutla2021mauve}, which quantifies the distributional gap between
generated and reference text.

\section{Results}
\label{sec:results}

\subsection{Efficiency: the state-space cache is constant in length}
\label{sec:res:eff}
\cref{fig:efficiency} and \cref{tab:efficiency} report per-step latency, peak memory, and decode throughput against context length at batch~1. The Mamba block cache delivers exactly $\mathcal{O}(1)$ complexity across the whole $64$--$256$k-token range, i.e. \bmamba\ holds a \emph{flat}
$6.8$\,ms per step, $7.5$\,GB of memory, and $\sim$$296$\,tok/s, its
state never growing with the prefix. \battn\ instead pays the
$\mathcal{O}(L)$ KV-cache in every panel: from $64$ to $256$k tokens its latency climbs $5.5\!\to\!29$\,ms ($5.3\times$), its decoding throughput falls $346\!\to\!114$\,tok/s ($3\times$), and its memory balloons $6.5\!\to\!82$\,GB, essentially the whole GPU. By $256$k, \bmamba\ thus beats \battn\ on all three axes at once: \hllat{$4.3\times$} lower latency ($6.8$ vs.\ $29$\,ms), \hlmem{$11\times$} less memory ($7.5$ vs.\ $82$\,GB), and \hltput{$2.6\times$} higher throughput ($297$ vs.\ $114$\,tok/s). \bhybrid\ is the pragmatic sweet spot: with only $5$ of its $28$ layers attentional it tracks the attention curves at $5/28$ of the slope, so at $256$k it keeps full attention for global recall while still beating \battn\ by \hllat{$2.7\times$} on latency ($10.7$ vs.\ $29$\,ms), \hlmem{$3.8\times$} on memory ($21.5$ vs.\ $82$\,GB), and \hltput{$2.0\times$} on throughput ($231$ vs.\ $114$\,tok/s).

The constant footprint pays off most under batching as shown in \cref{fig:efficiency_batch}: because \bmamba's per-stream memory does not grow, many sequences fit on one accelerator, and at $256$k it sustains $1593$\,tok/s at batch~$8$ (a flat $\sim$$8$\,GB), whereas \battn's $\mathcal{O}(B\cdot L)$ KV-cache is already out of memory at batch~$2$ and capped at $114$\,tok/s; a \hltput{$\mathbf{14\times}$} aggregate-throughput advantage (\autoref{app:efficiency_batch}). This dominance is a long-context effect: below $\sim$$16$k the three backbones sit within a few percent (with \battn\ marginally ahead on its cheap short-context blocks), but where long-context decoding actually lives, at tens to hundreds of thousands of tokens, \bmamba\ and \bhybrid\ win on latency, memory, and throughput at once. 
\subsection{Long-context generalization: linear state extrapolates, attention collapses}
\label{sec:res:long}
The position-free recurrence that makes the SSM cache cheap also makes it length robust. The mechanism is the block-causal mask: regardless of block size, a
query still attends over the \emph{entire} clean prefix, so past $L{=}1024$ the
attention layers meet absolute RoPE positions and prefix lengths never seen in
training (an out-of-distribution regime), whereas the recurrent state carries no
positions and simply folds a longer prefix into the same fixed-size memory.
\cref{tab:niah} bears this out on NIAH passkey retrieval. Within the $1024$ training
length all three models retrieve near-perfectly; at $2\times$ ($L{=}2$k) \battn\
\emph{collapses} to $12\%$ while \bhybrid\ holds $53\%$ and \bmamba\ $76\%$, and by
$16\times$ ($L{=}16$k) \battn\ sits at $0\%$ while \bmamba\ still recovers $22\%$ of needles. \bmamba\ owns the highest floor at the extremes ($2\times$ and $16\times$) and \bhybrid\ the middle of the range ($4\times$, and $8\times$ with NTK-RoPE). 

\begin{table}[t]
\centering
\caption{\textbf{Attention collapses at $2\times$ train length; linear state
extrapolates to $8$--$16\times$.} Needle-in-a-haystack passkey retrieval accuracy
(\%) vs.\ context length $L$ (3B; trained at $L{=}1024$, so $L{\ge}2$k is
extrapolation). ``+NTK'' adds NTK-RoPE to the attention-bearing models; best per
length \colorbox{best}{shaded}. \battn\ drops to $12\%$ at $2\times$ and $0\%$ by
$8\times$, while \bmamba/\bhybrid\ still retrieve needles at $16\times$.}
\label{tab:niah}
\small
\begin{tabular}{lcccccc}
\toprule
\rowcolor{headcol}
\hdr $L$ (= $\times$train) & \hdr 512 & \hdr 1k\,($1\times$) & \hdr 2k\,($2\times$) & \hdr 4k\,($4\times$) & \hdr 8k\,($8\times$) & \hdr 16k\,($16\times$) \\
\midrule
\battn            & 100 & 100 & 12.3 & 1.7  & 0    & 0    \\
\battn\,+NTK      & 100 & 100 & 12.3 & 1.7  & 0    & 0    \\
\bhybrid          & 98.7 & 100 & 53.4 & \cellcolor{best}37.9 & 23.0 & 2.3 \\
\bhybrid\,+NTK    & 98.7 & 100 & 53.4 & \cellcolor{best}37.9 & \cellcolor{best}26.1 & 14.3 \\
\bmamba           & 99.0 & 99.0 & \cellcolor{best}75.7 & 32.3 & 23.6 & \cellcolor{best}22.3 \\
\bottomrule
\end{tabular}
\end{table}

\begin{figure}[t]
\centering
\includegraphics[width=\linewidth]{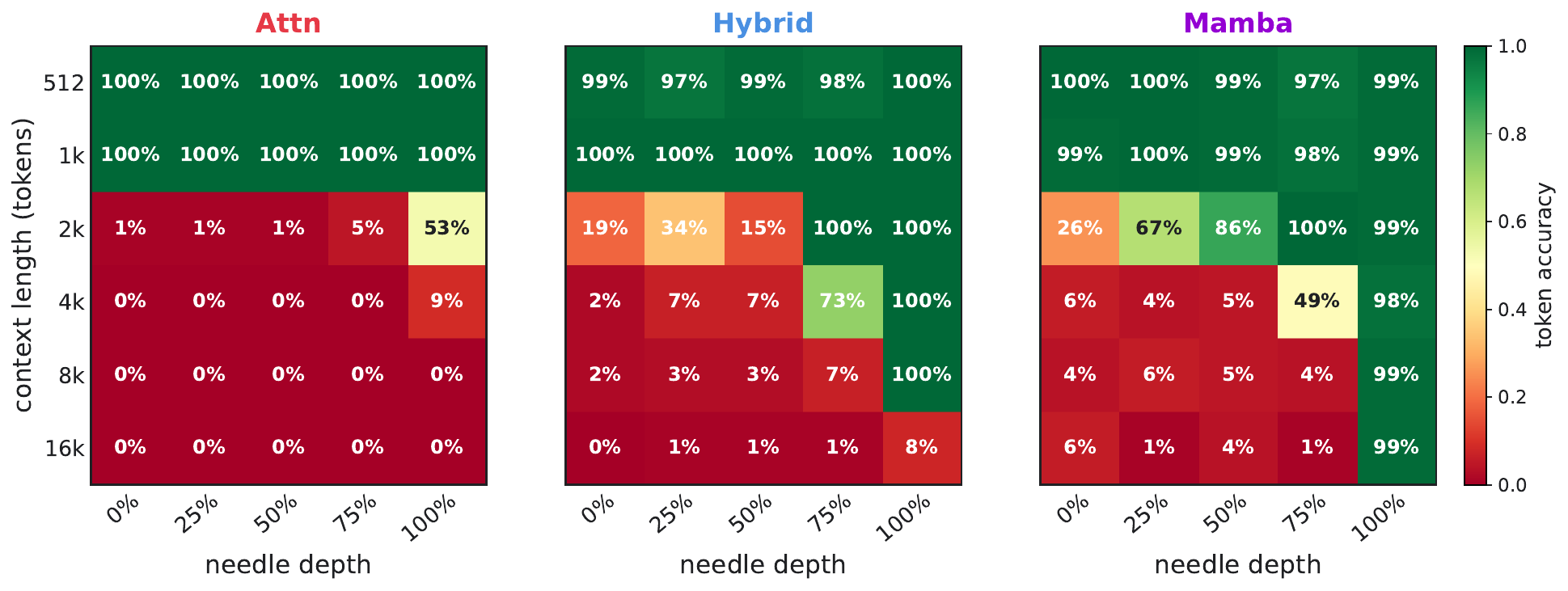}
\caption{\textbf{NIAH retrieval by needle depth $\times$ context length (3B).} Token
accuracy at five depths ($0\%$ = oldest, $100\%$ = most recent) across lengths.
\battn\ collapses almost uniformly past the $1024$-token training length;
\bmamba/\bhybrid\ degrade \emph{gracefully} and retain near-perfect retrieval of
\emph{recent} needles (rightmost column: \bmamba\ $99\%$ at $16\times$ training
length; per-depth numbers in \cref{app:niah}). This matches the ``local perception'' /
sliding-window behavior reported for diffusion LMs by LongLLaDA~\citep{liu2025longllada}, which appears here as a property of the linear-state backbone rather than of an approximate cache.}
\label{fig:niah_heatmap}
\vspace{-5pt}
\end{figure}

The same pattern holds on \emph{realistic} tasks (\cref{tab:longbench}). \bmamba\ is
the strongest backbone and barely degrades with length, holding essentially constant at
$10.2$--$11.2$ from $2$k to $16$k and ahead of \bhybrid\ ($10.8\!\to\!9.1$) at every
length. \battn, by contrast, falls by a third by $4$k ($7.1\!\to\!4.7$) and roughly
halves by $16$k ($\to\!3.6$), so the gap widens with context: at $16$k \bmamba\ nearly
triples and \bhybrid\ more than doubles \battn\ ($10.2$ and $9.1$ vs.\ $3.6$), just as
a position-agnostic recurrence that needs no length extrapolation predicts.

\begin{table}[t!]
\centering
\caption{\textbf{On realistic long-context tasks, linear state stays flat while
attention slides.} LongBench macro-average (\%) over 16 English+code tasks vs.\
length $L$ (3B, cached greedy decoding, middle-truncation, 20 examples/task; every
row $L{\ge}2$k is extrapolation from the $1024$ training length). ``+NTK'' $=$
NTK-RoPE on the attention-bearing models; best per length \colorbox{best}{shaded}.
\bmamba\ holds $10.2$--$11.2$ throughout; \battn\ more than halves by $4$k and
reaches $3.6$ at $16$k.}
\label{tab:longbench}
\small
\begin{tabular}{lccccc}
\toprule
\rowcolor{headcol}
\hdr $L$ & \hdr \battn & \hdr \battn\,+NTK & \hdr \bhybrid & \hdr \bhybrid\,+NTK & \hdr \bmamba \\
\midrule
2k  & 7.07 & 7.07 & 10.83 & 10.83 & \cellcolor{best}11.20 \\
4k  & 4.65 & 4.65 & 10.39 & 10.39 & \cellcolor{best}10.45 \\
8k  & 3.96 & 4.66 & 9.30 & 9.83 & \cellcolor{best}10.46 \\
16k & 3.64 & 4.75 & 9.13 & 8.10 & \cellcolor{best}10.18 \\
\bottomrule
\end{tabular}
\vspace{-10pt}
\end{table}
\begin{table}[t!]
\centering
\caption{\textbf{Constant-memory decoding at no downstream cost.} Accuracy on eight
common-sense/reasoning tasks (3B) via the block-diffusion likelihood harness
(block-decomposed MC-ELBO, 128 samples, CFG $0.5$); \textbf{Avg} is the mean,
best \colorbox{best}{shaded}. \battn\ and \bhybrid\ tie ($0.434$/$0.432$) and
\bmamba\ trails by $\sim$1 point.}
\label{tab:downstream}
\small
\setlength{\tabcolsep}{10pt}
\resizebox{\linewidth}{!}{%
\begin{tabular}{lcccccccc|c}
\toprule
\rowcolor{headcol}
\hdr Model & \hdr PIQA & \hdr ARC-c & \hdr HSwag & \hdr OBQA & \hdr BoolQ & \hdr SciQ & \hdr CSQA & \hdr SIQA & \hdr Macro \\
\midrule
\battn   & 0.655 & 0.226 & 0.347 & 0.210 & 0.616 & 0.829 & 0.188 & 0.400 & \cellcolor{best}\textbf{0.434} \\
\bhybrid & 0.663 & 0.224 & 0.339 & 0.214 & 0.621 & 0.815 & 0.187 & 0.392 & 0.432 \\
\bmamba  & 0.658 & 0.230 & 0.334 & 0.194 & 0.581 & 0.778 & 0.198 & 0.392 & 0.421 \\
\bottomrule
\end{tabular}%
}
\vspace{-10pt}
\end{table}
\begin{table}[t!]
\centering
\caption{\small \textbf{One dial, $S$, trades quality for speed.} Per-block denoising
budget $S$ (3B, cached generation, $64\to128$-token continuations, 512 samples):
larger $S$ monotonically lowers Gen-PPL and broadly raises MAUVE, at $5$--$6\times$
lower throughput (best per column \colorbox{best}{shaded}). MAUVE ($k$-means over
GPT-2-large features) is noisy for small base models, so Gen-PPL is the cleaner
signal.}
\label{tab:mauve}
\small
\setlength{\tabcolsep}{18pt}
\resizebox{\linewidth}{!}{%
\begin{tabular}{l ccc ccc}
\toprule
\rowcolor{headcol}
& \multicolumn{3}{c}{\hdr MAUVE $\uparrow$}
& \multicolumn{3}{c}{\hdr Gen-PPL $\downarrow$}\\
\cmidrule(lr){2-4}\cmidrule(lr){5-7}
\rowcolor{headcol}
\hdr $S$ & \hdr \battn & \hdr \bhybrid & \hdr \bmamba & \hdr \battn & \hdr \bhybrid & \hdr \bmamba \\
\midrule
4  & 0.055 & 0.050 & 0.043 & 15.26 & 17.14 & 18.50 \\
8  & 0.141 & 0.148 & 0.099 & 12.10 & 12.85 & 13.05 \\
16 & \cellcolor{best}0.236 & \cellcolor{best}0.289 & \cellcolor{best}0.287 & 9.22 & 9.06 & 9.50 \\
32 & 0.180 & 0.210 & 0.190 & \cellcolor{best}6.44 & \cellcolor{best}6.48 & \cellcolor{best}6.52 \\
\bottomrule
\end{tabular}%
}
\vspace{-15pt}
\end{table}

\subsection{Quality parity: efficiency at no cost}
\label{sec:res:quality}
The efficiency and long-context gains cost essentially nothing in quality. The three
models sit within $\sim$0.03 nats on validation NLL, with \bhybrid\ best
(\cref{tab:configs}), and on an eight-task downstream suite scored by the
block-diffusion likelihood harness (\cref{tab:downstream}) \battn\ and \bhybrid\ are
level (macro $0.434$ vs.\ $0.432$) while \bmamba\ trails by about a point. Thus the
hybrid delivers its $8$--$16\times$ length generalization and steeply reduced decode
memory at no measured quality cost, and the pure-SSM \bmamba\ gives up only $\sim$1
downstream point for fully constant-memory, length-independent inference.

\paragraph{Quality and throughput trade on a single dial.}
The per-block denoising budget $S$ trades quality for speed as shown in \cref{tab:mauve}. For all
three backbones, raising $S$ from $4$ to $32$ \emph{monotonically} lowers generative
perplexity ($15.3\to6.4$ for \battn, $17.1\to6.5$ for \bhybrid, $18.5\to6.5$ for
\bmamba) and broadly improves MAUVE, at $5$--$6\times$ lower throughput, so a deployment
can choose its operating point by turning one knob. One caveat: the absolute MAUVE
values are low and noisy (they peak near $S{=}16$ rather than rising strictly), because
these are small \emph{base} models producing short continuations, so Gen-PPL is the
cleaner quality signal here.

\subsection{The gains are architectural, not parameter count}
\label{sec:res:flops}
A natural worry is that Mamba's extra projection parameters, rather than its
linear-time mixing, drive the results. \cref{tab:flops} shows the advantage is a
property of the \emph{architecture}, not the parameter count. \bmamba\ and \bhybrid\ do
carry $13\%$ and $11\%$ \emph{more} parameters than \battn, but those parameters live in
$\mathcal{O}(d^2)$ projections that are independent of $L$ and add no long-context cost:
their forward FLOPs/token stay flat with context (\bmamba\ at $6.2$\,G for \emph{every}
length), whereas attention's parameter-free $\mathcal{O}(Ld)$ term grows to $24.2$\,G.
The three are comparable at the $1024$ training length, then diverge sharply: at
$L{=}64$k \bmamba\ and \bhybrid\ spend just $0.26\times$ and $0.39\times$ the attention
FLOPs. With quality already at parity (\cref{sec:res:quality}), the long-context win
rests on the linear-time mixer, not on the parameter budget: \bmamba\ and \bhybrid\ win
decisively at long context.

We calculate forward-pass FLOPs per token \emph{analytically} from the architecture
(two FLOPs per multiply--accumulate), summing every layer's mixer and gated FFN plus
the untied output head over the vocabulary; kernels are not profiled. Per token, an
attention layer costs $8d^2$ for the $Q,K,V,O$ projections plus $4Ld$ for the score
and value-aggregation matmuls, the \emph{only} $L$-dependent term. A bidirectional
Mamba-2 layer costs
$2\big[\,2d(2d_{\text{in}}{+}2d_{\text{state}}{+}n_h)+2d_{\text{in}}d+2d_{\text{in}}d_{\text{state}}+2d_{\text{conv}}(d_{\text{in}}{+}2d_{\text{state}})\,\big]$
with $d_{\text{in}}{=}d$ and $n_h$ Mamba heads (in/out projections, the SSD scan, and
the depthwise convolution, doubled for the two directions), all independent of $L$;
the output head adds $2dV$. Unlike the smaller-scale DiffuMamba hybrids~\citep{singh2026diffumamba}, the 3B Mamba/hybrid
variants keep the \emph{same} MLP as attention, so their extra mixer parameters are
not offset. \cref{tab:flops} evaluates this forward count for the 3B models
($d{=}2560$, $28$ layers, hybrid $=5$ attention $+\,23$ Mamba, gated FFN with hidden
size $7680$, $d_{\text{state}}{=}64$, expand $1$, $d_{\text{conv}}{=}4$, and an untied
vocabulary of $V{=}126\,464$) at $L{=}1024$ and $L{=}64$k; parameter counts are read
exactly from the trained checkpoints.

\begin{table}[t]
\centering
\caption{\textbf{More parameters, but the compute win is long-context.} Exact
parameters and analytic forward FLOPs/token for the 3B models (\cref{sec:res:flops}).
\bmamba\ and \bhybrid\ carry $11$--$13\%$ \emph{more} parameters than \battn, but these
live in $L$-independent $\mathcal{O}(d^2)$ projections that add no long-context cost:
per-token FLOPs are comparable at short context, while only the SSM cost stays flat with
length (\colorbox{best}{shaded}), so at $L{=}64$k \bmamba\ spends $0.26\times$ the
attention FLOPs while \battn's $\mathcal{O}(Ld)$ term dominates.}
\label{tab:flops}
\small
\begin{tabular}{lccc}
\toprule
\rowcolor{headcol}
\hdr Model & \hdr Params & \hdr FLOPs/tok ($L{=}1024$) & \hdr FLOPs/tok ($L{=}64$k) \\
\midrule
\battn   & 3.033\,B          & 5.712\,G & 24.209\,G \\
\bhybrid & 3.360\,B (+11\%)  & 6.139\,G & 9.442\,G \\
\bmamba  & 3.431\,B (+13\%)  & 6.232\,G & \cellcolor{best}6.232\,G \\
\bottomrule
\end{tabular}
\vspace{-10pt}
\end{table}

\section{Conclusion}
\label{sec:conclusion}
We pretrained attention, bidirectional-Mamba, and hybrid block-diffusion denoisers at
3B parameters and $300$B tokens under one block-causal frontier objective, and decoded
all three through a single, \emph{exact} cached interface whose state-space variant is
$\mathcal{O}(1)$ in sequence length. Concurrent small-scale work~\citep{chaturvedi2026training}
established that such a cache is possible; our results show what it delivers at scale.
At 256k tokens \bmamba\ decodes at a flat $6.8$\,ms/step and $7.5$\,GB where \battn\
needs $29$\,ms/step and $82$\,GB, and because that per-stream footprint never grows it
scales with batch to $14\times$ the aggregate throughput. The same linear-state bias
makes decoding length-robust: \bmamba\ and \bhybrid\ retrieve needles and answer
LongBench queries at $8$--$16\times$ their training length while \battn's retrieval
collapses at $2\times$, and the hybrid matches \battn\ on likelihood and downstream
accuracy, so these gains come at no measured quality cost.

\paragraph{Limitations and future work.}
Our models are trained at a $1024$-token context, so the long-context evaluations probe
\emph{extrapolation} rather than trained long-context ability, which explains the modest
absolute accuracies far past training length; closing this gap (via RoPE-scaling recipes
such as LongLLaDA/UltraLLaDA~\citep{liu2025longllada,he2025ultrallada}, or simply
training at longer contexts) is orthogonal to, and compatible with, our cache. Likewise,
we train with a single-frontier objective at a $1024$-token context, whereas
\citet{chaturvedi2026training} use an all-block objective at $8$k tokens; disentangling
the effect of objective and training length on cached long-context behavior is a natural
ablation we leave to future work. Natural
next steps are to combine our exact cache with confidence-aware parallel
decoding~\citep{fastdllm} inside the cached generator and to push block-cached hybrids to
larger scales. We believe constant-memory,
length-generalizing block decoding is a promising foundation for efficient
long-context diffusion language models.

\small
\bibliographystyle{abbrvnat}
\bibliography{refs}

\newpage
\appendix
\section{Training and model hyperparameters}
\label{app:hparams}
\Cref{tab:hparams} lists the full training and architecture configuration. The three
denoisers are identical except for the sequence mixer (\cref{tab:configs}); every other
setting is shared.

\begin{table}[t]
\centering
\caption{Full training and model hyperparameters, shared across \battn, \bhybrid, and
\bmamba\ (the sequence mixer is the only difference; \cref{tab:configs}).}
\label{tab:hparams}
\small
\begin{tabular}{ll}
\toprule
\rowcolor{headcol}
\hdr Hyperparameter & \hdr Value \\
\midrule
\multicolumn{2}{l}{\emph{Architecture}} \\
Model dimension $d_{\text{model}}$ & $2560$ \\
Layers & $28$ \\
Attention heads ($=$ KV heads) & $20$ \\
Head dimension & $128$ \\
MLP hidden size & $7680$ \\
Block size $G$ & $32$ \\
RoPE $\theta$ & $5\times10^{5}$ \\
Vocabulary (incl.\ mask) & $126\,464$ \\
Mixer (\battn\,/\,\bhybrid\,/\,\bmamba) & $28$ attn / $5$ attn $+$ $23$ Mamba-2 / $28$ Mamba-2 \\
\midrule
\multicolumn{2}{l}{\emph{Data}} \\
Corpus & Nemotron-CC~\citep{su2024nemotron} \\
Tokenizer & LLaDA tokenizer~\citep{llada} \\
Sequence packing & packed, no padding \\
\midrule
\multicolumn{2}{l}{\emph{Objective and optimization}} \\
Training objective & single-frontier block diffusion \\
Loss weighting & uniform \\
Optimizer & AdamW ($\beta_1{=}0.9$, $\beta_2{=}0.999$, $\epsilon{=}10^{-8}$) \\
Peak learning rate & $1\times10^{-4}$ \\
LR schedule & cosine decay \\
Warmup & $5\%$ of steps \\
Weight decay & $0.1$ \\
Gradient-norm clip & $1.0$ \\
Precision & \texttt{bf16} \\
\midrule
\multicolumn{2}{l}{\emph{Batching and token budget}} \\
Sequence length (training context) & $1024$ \\
Per-device batch & $16$ \\
Global batch & $4096$ sequences \\
Tokens per step & $1024\times4096\approx4.2$M \\
Training steps & $71\,525$ \\
Total tokens & $\approx$$300$B \\
\midrule
\multicolumn{2}{l}{\emph{Systems and checkpointing}} \\
Parallelism & FSDP \\
Attention kernel & PyTorch SDPA \\
Gradient checkpointing & off \\
\texttt{torch.compile} & off \\
Evaluation interval & every $100$ steps ($\le$$8192$ samples) \\
Checkpoint interval & every $50$ steps \\
Model selection & best validation NLL \\
\bottomrule
\end{tabular}
\end{table}

\section{Per-depth needle-in-a-haystack}
\label{app:niah}
\cref{tab:niah_depth} reports the retrieval accuracy behind \cref{fig:niah_heatmap}
by needle depth (0 = far from the end-of-sequence query, 1.0 = adjacent) at the two
lengths just past training length. All models show the recency/sliding-window
effect, but only the linear-state backbones retain any retrieval at $4\times$.

\begin{table}[h]
\centering
\caption{\textbf{The linear-state advantage is recency.} Per-depth NIAH retrieval
(\%) at $L{=}2$k ($2\times$) and $L{=}4$k ($4\times$) (3B; depth $0$ $=$ oldest,
$1.0$ $=$ adjacent to the query). Every backbone peaks on \emph{recent} needles, but
only \bmamba/\bhybrid\ keep retrieving older ones past training length; \battn\ is
near-zero at all but the last depth.}
\label{tab:niah_depth}
\small
\begin{tabular}{ll ccccc}
\toprule
\rowcolor{headcol}
\hdr $L$ & \hdr Model & \hdr depth 0.0 & \hdr 0.25 & \hdr 0.5 & \hdr 0.75 & \hdr 1.0 \\
\midrule
\multirow{3}{*}{2k}
 & \battn   & 1  & 1  & 1  & 5   & 53  \\
 & \bhybrid & 19 & 34 & 15 & 100 & 100 \\
 & \bmamba  & 26 & 67 & 86 & 100 & 99  \\
\midrule
\multirow{3}{*}{4k}
 & \battn   & 0 & 0 & 0 & 0  & 9   \\
 & \bhybrid & 2 & 7 & 7 & 73 & 100 \\
 & \bmamba  & 6 & 4 & 5 & 49 & 98  \\
\bottomrule
\end{tabular}
\end{table}

\section{Full batch-size efficiency sweep}
\label{app:efficiency_batch}
\begin{figure}[h]
\centering
\includegraphics[width=\linewidth]{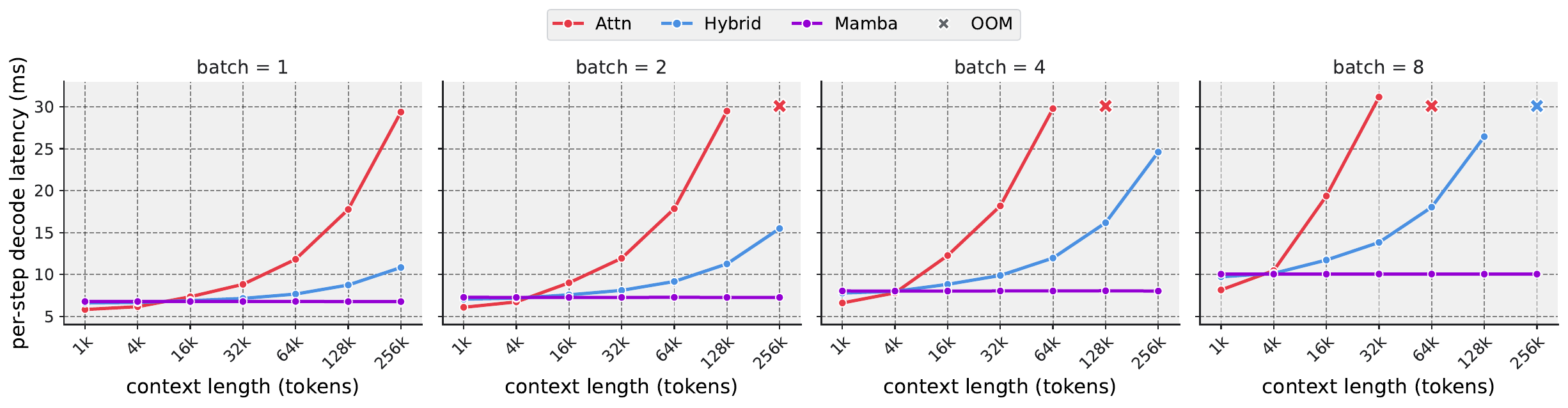}\\[3pt]
\includegraphics[width=\linewidth]{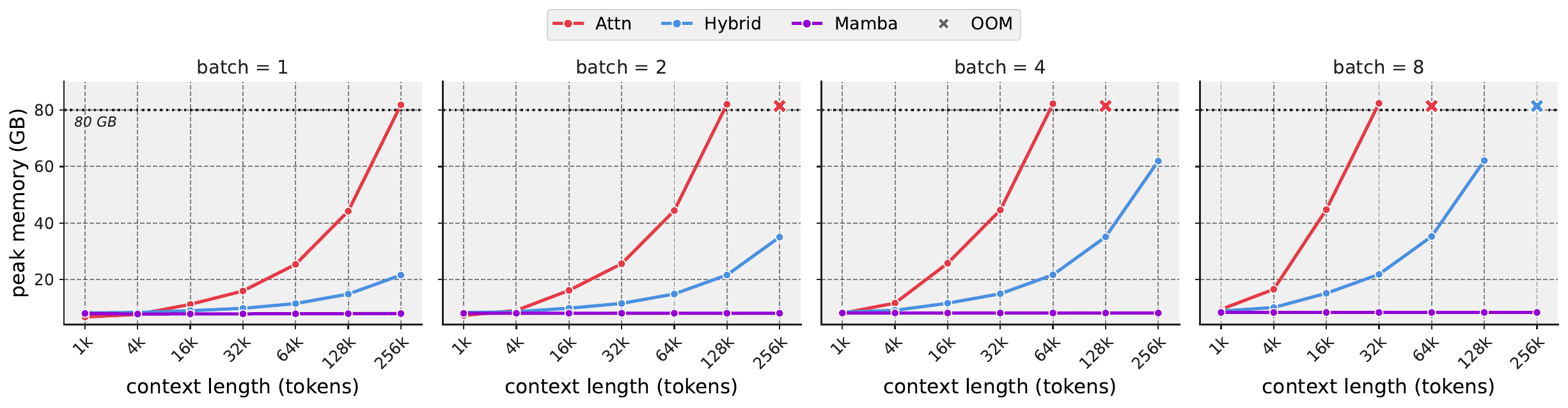}\\[3pt]
\includegraphics[width=\linewidth]{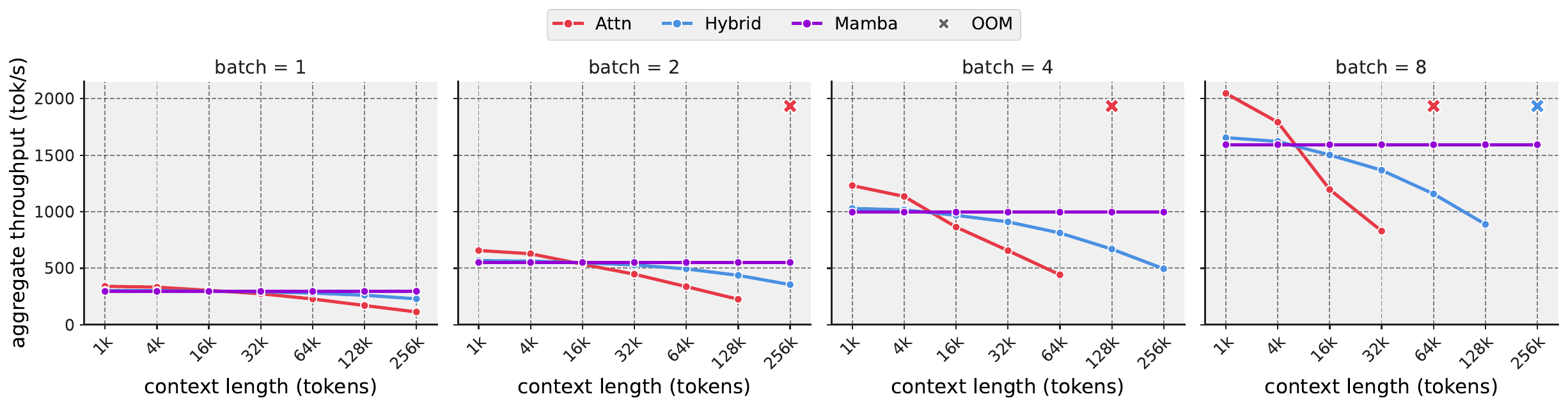}
\caption{\textbf{Efficiency vs.\ context length across batch sizes (3B, single
H100 80\,GB, \texttt{bf16}, cudagraph).} Top to bottom: per-step decode latency, peak
memory, aggregate decode throughput ($S{=}16$); within each row the four panels are
batch~$1/2/4/8$ (left to right). \bmamba\ stays $\sim$8\,GB at every (batch, length);
\battn's last-feasible length shrinks with batch (256k@B1, 128k@B2, 64k@B4, 32k@B8),
marked \texttt{OOM} ($\times$) beyond.}
\label{fig:efficiency_batch}
\end{figure}

\cref{fig:efficiency} plots batch~1 (exact numbers in \cref{tab:efficiency});
\cref{fig:efficiency_batch} gives the full
sweep over batch sizes $1/2/4/8$. \bmamba\ throughput \emph{scales with batch and
stays flat in length} (1593\,tok/s at batch~8 for \emph{every} length to 256k,
$\sim$$5.4\times$ the batch-1 rate, at a constant $\sim$8\,GB), while \battn's
$\mathcal{O}(B\!\cdot\!L)$ KV-cache exhausts the H100 (\texttt{OOM}) at
progressively shorter lengths as batch grows (last feasible: 256k@B1, 128k@B2,
64k@B4, 32k@B8). At 256k, \bmamba\ at batch~8 delivers 1593\,tok/s while \battn\
cannot run even batch~2; its best feasible aggregate is batch~1 at 114\,tok/s, the
$\sim$$14\times$ gap quoted in \cref{sec:res:eff}.

\begin{table}[h]
\centering
\caption{\textbf{Flat vs.\ $\mathcal{O}(L)$, in numbers.} Batch-1 per-step latency
(ms), peak memory (GB), and end-to-end decode throughput (tok/s, $S{=}16$) vs.\
context length (3B, H100 80\,GB, cudagraph), the exact values behind
\cref{fig:efficiency}. \bmamba\ is constant ($\sim$$6.8$\,ms, $\sim$7.5\,GB) out to
256k while \battn\ grows to $29$\,ms and $82$\,GB; best per length
\colorbox{best}{shaded}.}
\label{tab:efficiency}
\small
\setlength{\tabcolsep}{8pt}
\resizebox{\linewidth}{!}{%
\begin{tabular}{l ccc ccc ccc}
\toprule
\rowcolor{headcol}
& \multicolumn{3}{c}{\hdr Latency (ms) $\downarrow$}
& \multicolumn{3}{c}{\hdr Peak mem (GB) $\downarrow$}
& \multicolumn{3}{c}{\hdr Gen.\ tput (tok/s) $\uparrow$} \\
\cmidrule(lr){2-4}\cmidrule(lr){5-7}\cmidrule(lr){8-10}
\rowcolor{headcol}
\hdr $L$ & \hdr \battn & \hdr \bhybrid & \hdr \bmamba & \hdr \battn & \hdr \bhybrid & \hdr \bmamba & \hdr \battn & \hdr \bhybrid & \hdr \bmamba \\
\midrule
64     & \cellcolor{best}5.55 & 6.53 & 6.77 & \cellcolor{best}6.46 & 7.80 & 7.59 & \cellcolor{best}346 & 305 & 296 \\
4k     & \cellcolor{best}6.12 & 6.62 & 6.75 & 7.69 & 8.08 & \cellcolor{best}7.41 & \cellcolor{best}336 & 303 & 296 \\
16k    & 7.28 & 6.84 & \cellcolor{best}6.74 & 11.24 & 8.74 & \cellcolor{best}7.44 & \cellcolor{best}307 & 299 & 296 \\
64k    & 11.79 & 7.65 & \cellcolor{best}6.72 & 25.40 & 11.32 & \cellcolor{best}7.51 & 229 & 282 & \cellcolor{best}296 \\
128k   & 17.70 & 8.66 & \cellcolor{best}6.72 & 44.26 & 14.74 & \cellcolor{best}7.58 & 171 & 263 & \cellcolor{best}296 \\
256k   & 29.31 & 10.74 & \cellcolor{best}6.76 & 81.91 & 21.52 & \cellcolor{best}7.65 & 114 & 231 & \cellcolor{best}297 \\
\bottomrule
\end{tabular}%
}
\end{table}

\end{document}